\documentclass[runningheads]{llncs}
\usepackage[T1]{fontenc}
\usepackage{graphicx}
\usepackage{verbatim}
\usepackage{graphicx}
\usepackage[dvipsnames]{xcolor}

\usepackage{float}

\usepackage{amsmath, amssymb}
\usepackage{enumitem}
\usepackage{multirow}
\usepackage{hhline}
\usepackage{bm}
\usepackage[numbers,sort&compress]{natbib}
\usepackage[table]{xcolor}
\usepackage{url}
\usepackage{booktabs}
\usepackage{color}
\usepackage{caption}
\definecolor{shadecolor}{gray}{0.85}
\usepackage{etoolbox} 
\newtoggle{comments}
\newtoggle{anon}
\toggletrue{comments}
\toggletrue{anon}
\usepackage{makecell}
\usepackage[
n,
operators,
advantage,
sets,
adversary,
landau,
probability,
notions,
ff,
mm,
primitives,
events,
complexity,
asymptotics,
keys]{cryptocode}
\usepackage{listings}
\usepackage{xcolor}

\usepackage{enumitem}
\usepackage{stmaryrd}

\colorlet{punct}{red!60!black}
\definecolor{background}{HTML}{EEEEEE}
\definecolor{delim}{RGB}{20,105,176}
\colorlet{numb}{magenta!60!black}

\lstdefinelanguage{json}{
    basicstyle=\normalfont\ttfamily\scriptsize,
    showstringspaces=false,
    breaklines=true,
    frame=lines,
    backgroundcolor=\color{background},
    literate=
     *{0}{{{\color{numb}0}}}{1}
      {1}{{{\color{numb}1}}}{1}
      {2}{{{\color{numb}2}}}{1}
      {3}{{{\color{numb}3}}}{1}
      {4}{{{\color{numb}4}}}{1}
      {5}{{{\color{numb}5}}}{1}
      {6}{{{\color{numb}6}}}{1}
      {7}{{{\color{numb}7}}}{1}
      {8}{{{\color{numb}8}}}{1}
      {9}{{{\color{numb}9}}}{1}
      {:}{{{\color{punct}{:}}}}{1}
      {,}{{{\color{punct}{,}}}}{1}
      {\{}{{{\color{delim}{\{}}}}{1}
      {\}}{{{\color{delim}{\}}}}}{1}
      {[}{{{\color{delim}{[}}}}{1}
      {]}{{{\color{delim}{]}}}}{1},
}

\iftoggle{comments}{%
    \newcommand{\ivan}[2][]{\todo[color=red!30,#1]{\textsf{Ivan:} #2}}
  \newcommand{\darian}[2][]{\todo[color=blue!30,#1]{\textsf{Darian:} #2}}
  \newcommand{\minyi}[2][]{\todo[color=green!30,#1]{\textsf{Minyi:} #2}}
}{%
  \newcommand{\ivan}[2][]{}
  \newcommand{\darian}[2][]{}
  \newcommand{\minyi}[2][]{}
}

\usepackage[utf8]{inputenc}

\makeatletter
\newcommand{\linebreakand}{%
  \end{@IEEEauthorhalign}
  \hfill\mbox{}\par
  \mbox{}\hfill\begin{@IEEEauthorhalign}
}
\makeatother

\usepackage{txfonts}

\usepackage{mathtools}

\usepackage{csquotes}

\usepackage{algorithm}
\usepackage[noend]{algpseudocode}

\usepackage{xcolor}
\usepackage{tikz}
\usetikzlibrary{fit,calc}
\usetikzlibrary{tikzmark}

\colorlet{mypink}{red!40}
\colorlet{myblue}{cyan!60}
\colorlet{mygreen}{YellowGreen}
\colorlet{myorange}{YellowOrange}

\usepackage{hyperref}

\newcommand{\protocol}{\textsc{OSR}}
\begin{document}
\title{\protocol{}: Output Space Redistribution for Adaptive Label Removal in Classification Models}
\titlerunning{OSR for Adaptive Label Removal in Classification Models}
%




\author{Minyi PENG,
Darian Gunamardi, 
Ivan Tjuawinata, 
Yongsen Zheng, \\
Kwok-Yan Lam}

\authorrunning{M. Peng \textit{et al.}}
%
\institute{Nanyang Technological University, 50 Nanyang Ave, Singapore, 639798} 
%
\maketitle              
\begin{abstract}
Label removal occurs frequently in classification systems with evolving taxonomies, where categories must be dynamically updated or eliminated. To accommodate such changes, classification models must adapt accordingly. Existing solutions, broadly categorized as retraining-based and feature-space-adjustment-based, share common limitations despite their variations, including reliance on access to original data, substantial computational and storage costs, inconsistent results, poor scalability, and degradation of model utility. To address this, we propose a novel approach that leverages statistical redistribution in the output space to approximate the post-removal confidence vectors of a retrained model. Applicable as a modular output filter, our method bypasses the burden of feature-space adjustments or loss-function convergence, alleviating scalability limitations. Furthermore, by requiring only existing labels and prior output confidences, the method potentially mitigates privacy concerns inherent to data-dependent solutions. Extensive experiments demonstrate competitive performance against full retraining, with improvements in computational efficiency and privacy preservation across several classification tasks.

\end{abstract}

\section{Introduction}

Label removal during taxonomy or catalog revisions is a frequent occurrence in classification systems. Such removals are often formalized as category \emph{deprecation} or \emph{retirement}—for example, in e-commerce taxonomies, categories periodically merge, split, or discontinue as markets evolve, requiring downstream classifiers to retrain or adjust~\cite{ecommerce2025recat}.
Similarly, in financial industry classification systems such as the North American Industry Classification System (NAICS), codes are periodically merged, split, or redefined on multi-year revision cycles, necessitating downstream analyses and models to migrate away from retired codes~\cite{statcan2022naics}. To ensure alignment with evolving data, classification models trained on affected labels must be updated accordingly. Unlike relational databases, which offer relatively accessible information and diverse update mechanisms, no analogous solution exists for classification models.



Currently, there are two representative families of label-removal methods: retrain-ing-based label-removal methods \cite{SISA, yan2022arcane, chen2022grapheraser, chen2022receraser, aldaghri2021coded} and feature-space-adjustment-based label-removal algorithms \cite{influence_unlearning,weight_correction, hessian_forgetting, reweighting, nature_feature}. Retraining-based methods aim to make models forget targeted data and remove the associated information from the model. Compared to a fully retrained model that excludes the eliminated data, these methods seek to replicate its performance using fast, practical mechanisms that selectively erase the target information without full retraining. These methods typically require thorough removal and yield a new model with updated weights and outputs. Building on this, feature-space-adjustment-based algorithms focus on tracing parameters back to their last training state to retrieve the unlearned outputs, thereby bypassing full retraining and training-phase modifications. They require some access to the original training data or even its intermediary representations such as feature-space or hidden-layer activations, and the frequent involvement of Hessian-matrix computations renders these methods expensive and difficult to scale to large and complex models.

Despite their effectiveness, most existing methods suffer from several major limitations: a) Both retraining-based and feature-space-adjustment-based label-removal algorithms often require partial of full retraining, which often yields inconsistent behavior across runs and settings. More critically, feature space adjustment is highly prone to catastrophic consequences, thereby destabilizing performance. In contrast, our proposed method does not require access to the original training data, retraining, or any modifications to the model parameters or architecture. As a result, the original model is largely preserved while achieving substantial reductions in training time and computational complexity; b) These methods are not model-agnostic solutions applicable to all classifier models; instead, they are designed and optimized for specific classifier architectures and training settings. Consequently, their generalization across different classifiers is limited, and directly transferring them to other models typically requires additional adaptation, parameter tuning, and implementation refactoring, reducing universality and deployment flexibility. This lack of model-agnostic applicability imposes substantial customization and integration costs when deploying across classifiers.

To address these challenges and offer a more practical solution for model adaptation to label removal conditions, we propose a novel approximation approach that entirely avoids retraining, parameter or loss function modification, and direct access to the original training data. The proposed method achieves an efficient, consistent, and scalable removal effect directly on model output. By studying class-wise differences in the forward training phase, we approximate the training process, reverse its logic, and apply a filter at the output layer to adjust model outputs instantly. This filter operates in two steps, projection of the original output confidence vectors to remove the impact of deprecated labels, followed by redistribution of the remaining confidences to erase residual label influence while preserving model utility and post-removal performance. To validate the effectiveness of our method, experiments are conducted across small to large-scale vision models under single and multi-class removal settings. Results are directly compared against the gold standard of full retraining with detailed evaluations through statistical assessment of post-removal model performance. Extensive comparisons with leading parameter-based and loss function-based approaches further demonstrate the efficacy, consistency, and scalability of our method.

In summary, the main contributions of this work are as follows:
\begin{enumerate}
    \item We propose a novel training-free model approximation approach for label removal that operates directly in the output space, requiring no retraining, parameter or loss function modification, and scales efficiently across both single and multi-class removal requests.
    \item This approach approximates classification model label removal through boundary organization in softmax logits, employing a two-step filter of projection and redistribution to guarantee consistency and post-removal model utility, while requiring only lightweight label-level statistics in place of direct access to original training data or intermediary representations.
    \item Extensive experiments across four datasets demonstrate the effectiveness of the proposed method in adapting classification models to label removal conditions.
\end{enumerate}

\section{Related Work}
\subsection{Output Space Re-Distribution}

Output space re-distribution treats the softmax output itself, rather than the model's parameters, as the site of intervention, redistributing an existing class's probability mass among the remaining known classes without retraining the underlying representation. This connects to two lines of prior work that also treat a model's output distribution as a site for principled redistribution, though for different ends. In open-set recognition, OpenMax~\cite{bendale2016openmax} replaces the softmax layer with a calibrated alternative that redistributes probability mass toward a newly introduced ``unknown'' class whenever an input looks sufficiently unlike anything seen during training. OpenMax and our filter therefore share a mechanism---post-hoc redistribution of an existing model's output rather than a newly learned decision function---but the direction is reversed: OpenMax \emph{expands} the output space to accommodate a class the model never modeled, whereas our method \emph{contracts} it, redistributing an existing, well-modeled class's mass among the remaining known classes to approximate the confidence vector of a model that had never been trained on that class. 

Output space re-distribution shares a closer conceptual parallel with knowledge distillation, which similarly treats a model's full output distribution, not just the top prediction, as carrying transferable structure between classes~\cite{zhao2022decoupled}. Our method relies on the same intuition: the relative probabilities a model assigns to non-target classes encode inter-class structure, and the way a model redistributes a removed class's mass across the remaining classes draws on that same structure. The two approaches differ in how it is used: distillation transfers it by training a separate student model to reproduce a teacher's output behavior, whereas our method acts directly on the original model's output probabilities, without modifying parameters or requiring retraining.

\subsection{Label Removal in Class Unlearning}
Label removal through class unlearning targets the complete removal of one or more classes' influence from a trained model, falling under the broader scope of machine unlearning, which emphasizes complete deletion of targeted information from a model's parameters~\cite{acm_MU_survey}. The most common approach modifies parameters directly: Warnecke et al.~\cite{MU_features_labels} propose an influence-function-based method that estimates the training set's influence on parameters and applies a closed-form update, requiring access to the original training data and convexity/continuity assumptions on the loss. Related approaches similarly achieve label removal via closed-form solutions~\cite{suriyakumar2022algorithms} or through fine-tuning strategies that accelerate model updates~\cite{li2024fast}, sharing the common trait that removal is achieved by modifying the model's parameters, whether through direct weight updates, influence estimation, or gradient-based fine-tuning. 

Our method belongs instead to a line of work that shifts focus away from the parameter set toward the output layer. The most closely related is linear filtration, proposed by Baumhauer et al.~\cite{logit_transformation}: for classification models of the form $\sigma\left(\mathbf{W} \cdot f(\mathbf{x})\right)$ where $f(\mathbf{x})$ is a feature extractor and $\mathbf{W}$ is the final linear layer preceding softmax $\sigma$, they derive a closed-form linear transformation of $\mathbf{W}$ that removes the discriminative information associated with the deleted class, framing label removal as model sanitization under a black-box threat model with reduced cost relative to retraining. Both linear filtration and our method target class-wide label removal, avoid gradient-based retraining, and are motivated by computational efficiency. However, they differ in where the adjustment is made: linear filtration modifies the model's final linear layer directly, requiring the model to take the specific form $\sigma\left(\mathbf{W} \cdot f(\mathbf{x})\right)$ and requiring access to $\mathbf{W}$ itself. Our method instead never inspects or modifies any layer, operating entirely on the already-computed output confidence vector and deriving its redistribution directly from the model's emitted probabilities rather than the weights that produced them. This also affects generality: linear filtration is defined only for models with a linear decision boundary in the extracted features, whereas our filter applies unchanged to any model that outputs a probability vector, linear or otherwise.

\section{Problem Statement}
In this work, we consider the problem of classification model adaption to label removal conditions, where a pretrained classification model must be updated to accommodate one or more label removal requests. The adapted model should retain both the knowledge of the remaining labels and the ability to apply that knowledge, while matching the performance of a newly trained model on the updated label set. The adaptation process must also be realistic, considering aspects such as instancy, efficiency, compatibility and the level of access availability to the model and data.

Let $f:\mathcal{X}\rightarrow[0,1]^{n}$ be a classification model that has been trained using a dataset $\mathcal{D}^T$ with labels in $\mathcal{L}=\{1,\cdots, n\}.$  Let $\mathcal{D}$ be the reference data for the label removal process. Suppose that the set of labels to be removed are $\mathcal{L}_{del}\subseteq \mathcal{L}.$ Define $\mathcal{L}_r=\mathcal{L}\setminus\mathcal{L}_{del}.$ We aim to develop a procedure $\Pi$ such that given $f,$ reference data $\mathcal{D},$ the removed labels $\mathcal{L}_{del},$ and an input $\mathbf{x},$ it produces the confidence vector of the adjusted model of length $n-1$ corresponding to labels in $\mathcal{L}_r$ with input $\mathbf{x}.$ Specifically, given $f, \mathcal{D}$ and $\mathcal{L}_{del}, \Pi$ produces a classification model $f':\mathcal{X}\rightarrow [0,1]^{n-|\mathcal{L}_r|}$ such that $f'$ approximates $f^\ast:\mathcal{X}\rightarrow [0,1]^{n-|\mathcal{L}_r|},$ a classifier model retrained from scratch using $\mathcal{D}_r\triangleq \{(\mathbf{x},y)\in \mathcal{D}^T:y\in \mathcal{L}_r\}.$ Here the measure of approximation is defined based on commonly used machine learning metrics including the accuracy difference and the KL divergence between the two models.

We assume that the pre-trained model $f$ outputs a confidence vector. Define $\hat{f}$ as a related function that outputs final predicted decision by taking the label corresponding to the highest confidence level of $f$. We further assume some performance guarantee requirements on the pre-trained model $f.$
We require that $f$ is sufficiently well trained to learn information regarding the training data. This is represented by the ability of $\hat{f}$ to predict the correct label with high enough accuracy. Furthermore, it can also be observed from the fact that for any input $x,$ if $\hat{f}(x)$ outputs the correct label, its corresponding confidence is sufficiently high. This requirement ensures the variance of the predicted confidence in $f$ to be small.
These constrain the variance of predicted confidence in $f$. By the assumptions we have made, we can assume that the confidence vectors output by $M^P$ for such data points have a smaller variance around their averages. This provides a sharper estimate of the projection space.

To approach the problem, we first consider how classification training is done in general. A classical classification model training uses training data with all possible $n$ labels simultaneously to optimize a function that outputs a real-valued vector of length $n$, containing probability of the data being from each of the $n$ labels. Here we note that under some smoothness and convexity assumptions of the loss function, a similar model can be obtained through the use of class incremental learning (CIL) (see for example \cite{ZWQ+24}) where labels are introduced sequentially, which is an implication of basic convex optimization theory. This also implies that models established by using CIL in different label introduction orders will converge to the same global minimum. One phenomenon that we need to consider in CIL is the catastrophic forgetting. This may happen when training using data with the next label without reconsidering the data from the previously learned labels. To avoid such phenomenon, CIL typically requires some information about the data from previously learned labels to influence the current iteration of training. In this work, we are particularly interested in the effect of introducing a new label in CIL from the perspective of the evolution of label boundary, which interpretation is formalized in the following section \ref{sec:forward_alg}.

\section{Methodology}

\begin{figure}[t]
    \centering
    \includegraphics[width=0.9\textwidth]{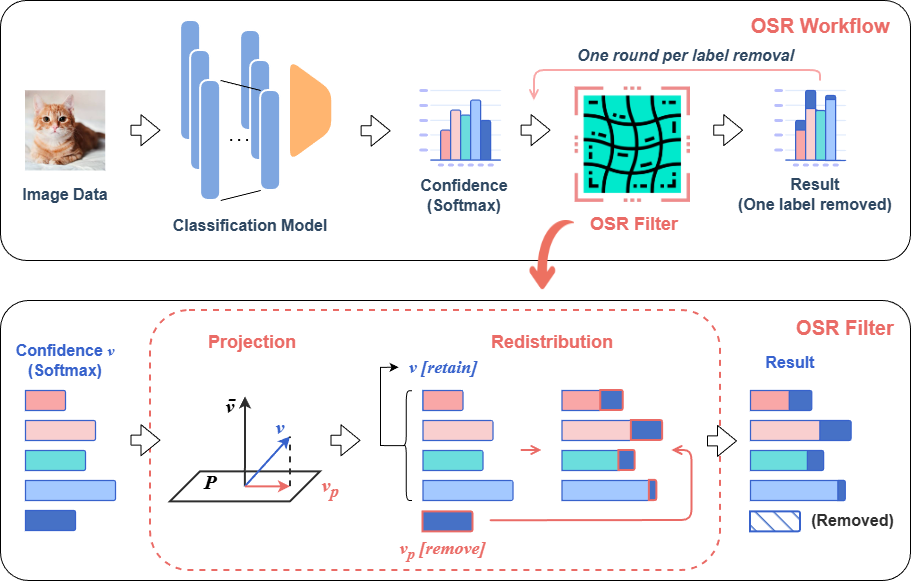}
    \caption{Overview of our OSR filter, which applies a post-hoc output-space redistribution comprising a projection step and a redistribution step. It removes the influence of the removed labels, reweights the remaining labels, and reassigns the residual probability mass associated with the removed labels, using a classifier trained on the updated label set as reference. This adjustment revises the original predicted confidence while preserving semantic consistency.}
    \label{fig:osr_visual}
\end{figure}

Our method treats adaptation to label removal as a boundary reorganization problem. Specifically, it considers how a new boundary expands during the training phase of classification, and how to reverse that process back to the previous boundary state. The following section \ref{sec:forward_alg} shows how boundary expansion occurs when a new label is learned. Section \ref{sec:OSR_alg} explains how the boundary shrinks to erase the label's effect and its residual influence, under the condition of a single label removal. Given the lightweight nature and low complexity of this procedure, as demonstrated in section \ref{sec:complexity}, the classification model handles multi-label removal through iteration of single-label removal. A visualization and workflow of the OSR procedure can be found in Fig. \ref{fig:osr_visual}.

\subsection{Boundary Expansion considered CIL}\label{sec:forward_alg}
We consider the following strategy of how CIL happens. For $i=2,\cdots, n,$ we start from $f_{i-1}:\mathcal{X}\rightarrow [0,1]^{i-1},$ trained using data points with labels belonging to $\{1,\cdots, i-1\}.$ For $j=1,\cdots, n,$ we denote by $\mathcal{D}_j$ the subset of $\mathcal{D}$ containing data points with label $j$ and $\mathcal{D}^{(c)}_j=\bigcup_{i=1}^j \mathcal{D}_j.$ At step $i,$ we consider the training that updates $f_{i-1}$ to $f_i.$ Intuitively, this is done by first embedding the output space of $f_{i-1}$ to $[0,1]^i,$ which is then trained further using $\mathcal{D}_i$ to obtain $f_i.$ This process can be summarized in the following 4 steps.

\begin{itemize}
    \item \textbf{Stage $1$.} For each $x\in \mathcal{X},$ denote $\mathbf{c}^{(i-1)}_x= f_{i-1}(x)$ as a confidence vector of length $i-1.$ 
    \item \textbf{Stage $2$.} Given $\mathbf{c}^{(i-1)}_x,$ we initialize a small constant $\alpha_0(x)>0,$ and normalize $\mathbf{c}^{(i-1)}_x$ to have sum $1-\alpha_0(x).$ This produces a new vector $((1-\alpha_0(x))\mathbf{c}^{(i-1)}_x\|\alpha_0(x))\in [0,1]^i $ of length $i$ with sum of entries being equal to $1$.
    \item \textbf{Stage $3$.} Train $f_i:\mathcal{X}\rightarrow [0,1]^i$ which is a classification model with $i$ labels using $\mathcal{D}_i$ as the train data set with further information obtained from $f_{i-1}(x)$ for $x\in \mathcal{D}_{i-1}^{(c)}$ to optimize the values of $\alpha_0(x)$ and $\mathbf{c}_x^{(i-1)}$. More specifically, we train $\alpha(x)$ and $\mathbf{c}_x^{(\mathtt{trans})}$ where $f_i(x)=((1-\alpha(x))\mathbf{c}^{(\mathtt{trans})}_x\|\alpha(x)),$ here $\alpha(x)$ is initialized as $\alpha_0(x)$ and $\mathbf{c}^{(\mathtt{trans})}_x$ is initialized as $\mathbf{c}^{(i-1)}_x.$
    \item \textbf{Stage $4$.} Once the training in stage $3$ is done, we have the final values of $\alpha(x)$ and $\mathbf{c}^{(\mathtt{trans})}_x$ which we denote by $\alpha_T(x)$ and $\mathbf{c}^{(T)}_x.$ We define $\mathbf{c}^{(i)}_{x}\triangleq f_i(x) = ((1-\alpha_T(x))\mathbf{c}^{(T)}_x\|\alpha_T(x)).$ 
\end{itemize}

\begin{figure}[h]
    \centering    \includegraphics[width=1.0\textwidth]{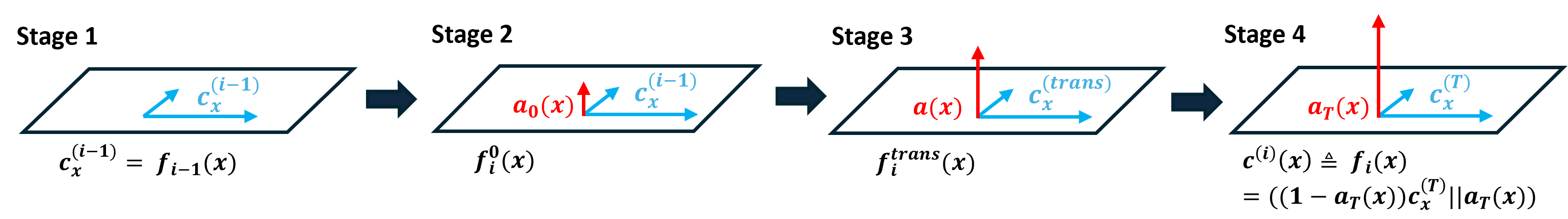}
    \caption{Transition Within the Output Space When Learning a New Label}
    \label{fig:learning_process}
\end{figure}

Visualization of the process is reflected in Figure \ref{fig:learning_process}. Note that for $x\in \mathcal{D}_i,$ assuming that $f_i$ has a sufficiently high accuracy for input $x,$ we can see $f_i(x)$ as approximately orthogonal to $f_{i-1}(x')$ for $x'\in \mathcal{D}^{(c)}_{i-1}.$ Hence, in order to recover $f_{i-1}(x')$ for $x'\in \mathcal{D}^{(c)}_{i-1},$ we may consider the projection of $f_i(x)$ to the vector space orthogonal to $f_i(x)$ for $x\in \mathcal{D}_i.$

\subsection{Reverse Output Semantics of One label}\label{sec:OSR_alg}

\begin{figure}[!t]
    \centering
    \includegraphics[width=0.9\textwidth]{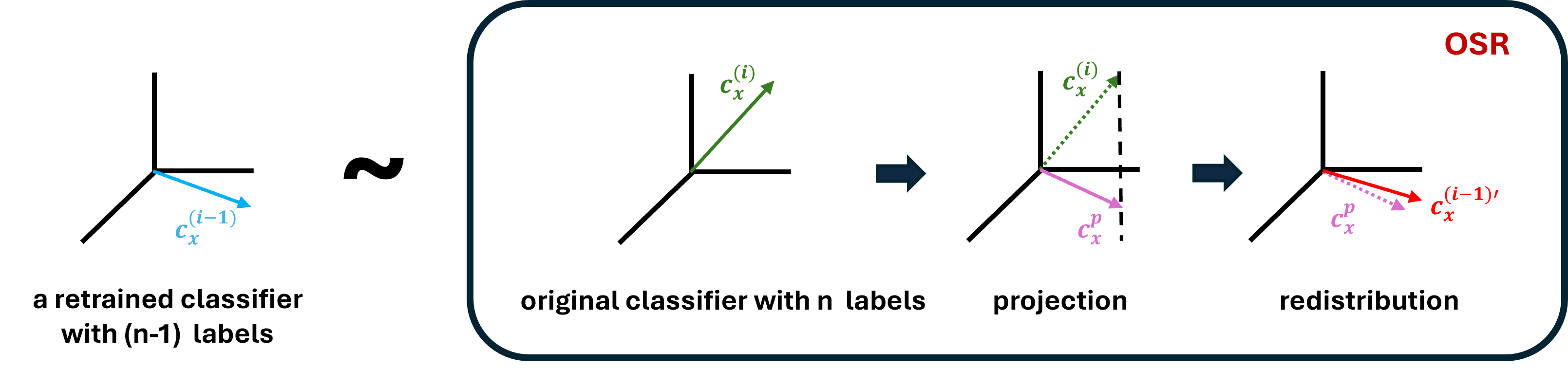}
    \caption{Reversing the Transition Within the Output Space to Remove An Existing Label}
    \label{fig: OSR}
\end{figure}

To recover the semantics of the classification model $f_{i-1}(x)$, we reverse the update from $f_i(x)$ to $f_{i-1}(x)$ in the output vector space. To achieve this, we adopt the learning transition described in Section \ref{sec:forward_alg} and model its reverse process, as illustrated in Fig. \ref{fig: OSR}. Starting from $f_i(x)$, which has output dimension $n$, we first construct a reference space $\mathcal{U}$ that approximates the output vector space of $f_{i-1}(x)$.

\begin{itemize}
    \item Projection with Reference Space $\mathcal{U}$ \\
    Given a group of test data $\mathcal{D}_i^{test}$ with to-be-removed label $i \in \mathcal{L}$, an average confidence vector $\bar{\mathbf{c}}_i$ is generated with respect to classification model predictions $f_i(x)$ where $x \in \mathcal{D}_i^{test}$. The null space of the average vector $\bar{\mathbf{c}}_i$ is computed with Gauss-Jordan elimination to get basis vectors $\{\mathbf{v}_{i,1},\cdots, \mathbf{v}_{i,n-1}\}\subseteq\mathbb{R}^n$ of solution space $\bar{\mathbf{c}}_i^T\cdot\mathbf{v}=0$. Further, Gram-Schmidt orthogonalisation is performed to obtain a set of orthonormal basis $\{\mathbf{q}_{i,1},\cdots, \mathbf{q}_{i,n-1}\}$ based on the intermediate basis $\{\mathbf{v}_{i,1},\cdots, \mathbf{v}_{i,n-1}\}$. This new basis $\{\mathbf{q}_{i,1},\cdots, \mathbf{q}_{i,n-1}\}$ constructs a matrix $A_i$ corresponding to the Reference Space $\mathcal{U}$ with its projection matrix $P_i = A_i \cdot A_i^T$. For any vector $v$ projected to the reference space $\mathcal{U}$, a projected vector should be obtained by $v^p = P_i \cdot v$.
\end{itemize}

Based on our assumption, the average confidence vector $\bar{\mathbf{c}}_i$ should capture the characteristics of the removed label $i$. The reference space $\mathcal{U}$ orthogonal to $\bar{\mathbf{c}}_i$ scrubs most information off from it. When the confidence vectors $\mathbf{c}_i$ output by classification model $f_i$ for such data points have a smaller variance around their averages, this should provide a sharper estimate of the reference space. This reference plane $\mathcal{U}$ indicated by matrix $A_i$ resembles the decision plane of $f_{i-1}(x)$, as shown in Stage 1 in Section \ref{sec:forward_alg}. For any confidence vector $\mathbf{c}_x$ output by classification model $f_i(x)$, we perform projection to the reference space to obtain its projected confidence $\mathbf{c}_x^p$. Here after, we split the confidence vector into scalar of the removed label $i$ and sub-vector of the remaining labels $\mathcal{L}\setminus \{i\}$. This is to emulate the concatenate vector $(1-\alpha_T(x))\mathbf{c}^{(T)}_x\|\alpha_T(x)$ as denoted in Stage 2 to 4 to further tackle them separately.

We first consider the scalar confidence score $c_x^p$ of the removed label $i$ from the projected vector $\mathbf{c}_x^p$. It is relatively small after projection and resembles $\alpha_0(x)$ defined in Stage 2. Consider the initialization of $\alpha_0(x)$, it is a small ratio coming from the confidence vector $\mathbf{c}^{(i-1)}_x$ output by classification model $f_{i-1}(x)$. Consider from another perspective of classification model $f_{i}(x)$, $c_x^p$ indicates the residual impact of the removed label $i$. Therefore, one more step is made to further distribute this residual probability $c_x^p$ back to the original semantic vector $\mathbf{c}^{(i-1)}_x$.

\begin{itemize}
    \item Redistribution of the Removed label Probability \\
    Given the same group of test data $\mathcal{D}_i^{test}$ that generates matrix $A_i$ corresponding to the Reference Space $\mathcal{U}$ above, all of their predictions $y = f_i(x)$ are projected to the space $\mathcal{U}$, returning a list of projected removed label confidences $y_x^p$. For every entry in the projected confidence $y_x^p$, we take its absolute value to form a new vector $z_x^p$ and further normalize it to have sum 1, denoted as vector $\tilde{z}_{x}^p$. By removing its $i$-th entry, it results a new list of vectors $\tilde{z}_{x,r}^p$. The distribution ratio vector $\bar{\mathbf{c}}_{i,r}$ for any residual probability $c_x^p$ is then obtained by averaging the list of $\tilde{z}_{x,r}^p$. 
    
    For any residual probability $c_x^p$, we have its removed result $c_x^P\cdot \bar{\mathbf{c}}_{i,r}$.
\end{itemize}

We then consider the sub-vector $\mathbf{c}_x^r$ of the confidence vector $\mathbf{c}_x$ of length $i-1$. It indicates the confidence values of the  remaining labels $\mathcal{L}\setminus \{i\}$. We reckon the output semantics of the remaining $i-1$ dimensions not undergoing an obvious change, thus the redistribution is about adjusting its ratio back to the portion of $\mathbf{c}_x^{i-1}$ in Stage 2.

\begin{itemize}
    \item Redistribution of the Remaining label Probability \\
    The proportion of the remaining label probability $\mathbf{c}_x^r$ in classification model $f_i(x)$ is denoted by $1-c_x^U$, where $c_x^U$ is the $i$-th entry of the predicted confidence vector $\mathbf{c}_x$ from classification model $f_i(x)$. When the output semantics returns to the initialized stage - Stage 2, the ratio of $\mathbf{c}_x^r$ turns back to $1-\alpha_0(x)$. This ratio is approximated by $1-c_x^p$, corresponding to the projected confidence value of the removed label. 

    For any remaining label probability $\mathbf{c}_x^r$, we have its redistributed result $\frac{1-c_x^P}{1-c_x^U}\cdot \mathbf{c}_x^r$.
\end{itemize}
The complete predicted result of reversing output semantics of a label is to add the separate result of its removed label $c_x^P\cdot \bar{\mathbf{c}}_{i,r}$ and the remaining labels $\frac{1-c_x^P}{1-c_x^U}\cdot \mathbf{c}_x^r$ together, normalize to have sum 1. With this entire operation, the reversed output semantics is expected to preserve the original distribution of classification model $f_i(x)$, while ensembles the predicted decisions of classification model $f_{i-1}(x)$. Details of the algorithm are provided in Algorithm \ref{alg:OSR}. 


\newcolumntype{R}[1]{>{\raggedleft\arraybackslash}p{#1}}

\begin{algorithm}[t]\label{sec:alg_OSR}
\caption{OSR $\mathbf{v}\leftarrow \mathtt{OSR}(f,i,\mathcal{D}_i,x)$
}
\label{alg:OSR}

\begin{algorithmic}[1]
\State Initiate $\bar{\mathbf{c}}_i\leftarrow \mathbf{0}_n$ where $\mathbf{0}_n$ is the zero vector of length $n;$
\For{$x\in \mathcal{D}_i$}
\State Calculate $y_x\leftarrow f(x);$
\State $\bar{\mathbf{c}}_i\leftarrow \bar{\mathbf{c}}_i+y_x;$
\EndFor
\State $\bar{\mathbf{c}}_i\leftarrow \frac{1}{|\mathcal{D}_i|}\bar{\mathbf{c}}_i;$
\State Perform Gauss-Jordan elimination to find $\{\mathbf{v}_{i,1},\cdots, \mathbf{v}_{i,n-1}\}\subseteq\mathbb{R}^n,$ the basis for the solution space of $\bar{\mathbf{c}}_i^T\cdot\mathbf{v}=0;$
\State Perform Gram-Schmidt Orthogonalization from $\{\mathbf{v}_{i,1},\cdots, \mathbf{v}_{i,n-1}\}$ to obtain an orthonormal basis $\{\mathbf{q}_{i,1},\cdots, \mathbf{q}_{i,n-1}\}$ of the solution space of $\bar{\mathbf{c}}_i^T\cdot \mathbf{v}=0;$
\State Define $A_i\in \mathbb{R}^{n\times n-1}$ with $\mathbf{q}_{i,j}^T$ as its $j$-th column and $P_i=A_i\cdot A_i^T$;
\For{$x\in \mathcal{D}_i$}
\State Calculate $y_x^p\leftarrow P_i\cdot y_x;$
\State Calculate $z_x^p,$ by taking the absolute value of each entry of $y_x^p$, and $\tilde{z}_x^p,$ obtained by normalizing $z_x^p$ to have sum $1;$
\State Construct $\tilde{z}_{x,r}^p$ from $\tilde{z}_x^p$ by removing its $i$-th entry;
\EndFor
\State Calculate $\bar{\mathbf{c}}_{i,r}\leftarrow \frac{1}{N_i} \sum_{x\in\mathcal{X}} \tilde{z}_{x,r}^p;$
\State Calculate $\mathbf{c}_x \leftarrow f(x).$ Denote by $c_x^u,$ its $i$-th entry and $\mathbf{c}_x^r,$  the vector obtained from $\mathbf{c}_x$ by removing its $i$-th entry;
\State Calculate $\mathbf{c}_x^P \leftarrow P_i\cdot \mathbf{c}_x$ and retrieve $c_x^P,$ its $i$-th entry;
\State Calculate $\mathbf{c}_1\leftarrow c_x^P\cdot \bar{\mathbf{c}}_{i,r};$
\State Calculate $\mathbf{c}_2\leftarrow \frac{1-c_x^P}{1-c_x^U}\cdot \mathbf{c}_x^r$ and $b\leftarrow c_x^U+1-c_x^P;$
\State Output $\frac{1}{b}\left(\mathbf{c}_1+\mathbf{c}_2\right);$
\end{algorithmic}
\end{algorithm}

\subsection{Asymptotic Complexity Analysis}\label{sec:complexity}

We analyze the asymptotic complexity of our proposed procedure. We assume complexity is measured in terms of the number of real number multiplication or division operations, and that real number addition has negligible complexity by comparison. We denote by $\mathcal{C}_{\mathtt{Div}}$ the complexity of a single real number division operation, measured in terms of the number of real number multiplications, and by $\mathcal{C}(f)$ the complexity of one call to $f$, likewise measured in terms of real number multiplications.  We further note that normalization and averaging can be done by first calculating $\frac{1}{X}$ where $X$ is the denominator before multiplying it to the vector being averaged or normalized. This allows the reduction of number of divisions (which is the more expensive operation) by replacing it with $1$ division operation and multiple multiplication operations. Table~\ref{tab:complexity} summarizes the complexity of each step in our Algorithm \ref{alg:OSR}.

\begin{table}[htbp]
\centering
\caption{Asymptotic complexity of steps in Algorithm \ref{alg:OSR}.}
\label{tab:complexity}
\begin{tabular}{@{}l|l@{}}
\toprule
\textbf{Step (of algorithm 1)} & \textbf{Complexity (in number of multiplication operations)} \\
\midrule
Steps $2$-- $4$ & $N_i \cdot \mathcal{C}(f)$ \\
Step $5$ & $\mathcal{C}_{\mathtt{Div}}+n$ \\
Step $6$--$7$ & $O(n^3)$ \\
Step $8$ & $O(n^3)$ \\
Steps $9$--$13$ & $(N_i+1)\mathcal{C}_{\mathtt{Div}}+O(N_in^2)$ \\
Step $14$ & $\mathcal{C}(f)$ \\
Step $15$ & $O(n^2)$ \\
Step $16$ & $n-1$ \\
Step $17$ & $\mathcal{C}_{\mathtt{Div}} + (n-1)$ \\
Step $18$ & $\mathcal{C}_{\mathtt{Div}}+(n-1)$ \\
\midrule
\textbf{Total} & $(N_i+1)\cdot\mathcal{C}(f)+(N_i+4)\cdot\mathcal{C}_{\mathtt{Div}}+O\left(n^3+N_in^2\right) \approx O(N_i)C(f)$ \\
\bottomrule
\end{tabular}
\end{table}

\begin{table}[t]
\centering
\caption{Comparison of accuracy metrics on CIFAR10, CIFAR100, and VGGFace-100 between base model (B), Retrained, (R), OSR (ours), UNSIR (\cite{TCMK23}), and NHLE-CWM (\cite{nature_feature}). Bolded entries show the best performance in each setting.} 
\label{tab:AM}
\begin{scriptsize}
\begin{tabular}{|p{0.1\linewidth}||p{0.12\linewidth}||p{0.09\linewidth}||p{0.07\linewidth}|p{0.07\linewidth}|p{0.07\linewidth}|p{0.07\linewidth}||p{0.07\linewidth}|p{0.07\linewidth}|p{0.07\linewidth}|p{0.07\linewidth}|}
\hline
\multirow{2}{*}{\textbf{Dataset}}&\multirow{2}{*}{\textbf{Model}}& \multirow{2}{*}{\textbf{Method}}&\multicolumn{4}{c||}{\textbf{Forget Classes}($\mathcal{A}_{\mathcal{R}}$) $\uparrow$}&\multicolumn{4}{c|}{\textbf{Forget Classes} ($\mathcal{A}_{\mathcal{F}}$) $\downarrow$}\\
\cline{4-11} 
&&&$\mathcal{F}_1$&$\mathcal{F}_2$&$\mathcal{F}_3$&$\mathcal{F}_4$&$\mathcal{F}_1$&$\mathcal{F}_2$&$\mathcal{F}_3$&$\mathcal{F}_4$\\
\hline
\hline
\multirow{10}{1cm}{CIFAR-10}&\multirow{5}{*}{ResNet18}&B&0.9296&0.9263&0.9173&0.9423&0.929&0.9425&0.9478&0.924\\
\cline{3-11}
&&R&0.9314&0.9325&0.9243&0.9823&\textbf{0}&\textbf{0}&\textbf{0}&\textbf{0}\\
\cline{3-11}
&&OSR&\textbf{0.9352}&\textbf{0.9386}&\textbf{0.9425}&\textbf{0.9953}&\textbf{0}&\textbf{0}&\textbf{0}&\textbf{0}\\
\cline{3-11}
&&\cite{TCMK23}&0.8891&0.8773&0.8242&0.9523&\textbf{0}&\textbf{0}&0.0005&0.0014\\
\cline{3-11}
&&\cite{nature_feature}&0.9044&0.9035&0.8448&0.9377&\textbf{0}&\textbf{0}&\textbf{0}&0.0047\\
\hhline{|~||==========}
&\multirow{5}{*}{AllCNN}&B&0.9181&0.9139&0.9062&0.9347&0.861&0.9065&0.9218&0.9029\\
\cline{3-11}
&&R&0.9187&0.9193&0.919&0.975&\textbf{0}&\textbf{0}&\textbf{0}&\textbf{0}\\
\cline{3-11}
&&OSR &\textbf{0.922}&\textbf{0.9284}&\textbf{0.9312}&\textbf{0.9943}&\textbf{0}&\textbf{0}&\textbf{0}&\textbf{0}\\
\cline{3-11}
&&\cite{TCMK23}&0.8737&0.8629&0.8353&0.917&0.004&0.0025&0.005&0.0007\\
\cline{3-11}
&&\cite{nature_feature}&0.892&0.906&0.751&0.882&\textbf{0}&\textbf{0}&0.0003&0.01529\\
\hline\hline
\multirow{10}{1cm}{CIFAR-100}&\multirow{5}{*}{ResNet18}&B&0.8171&0.8179&0.8167&0.8165&0.94&0.82&0.8208&0.8195\\
\cline{3-11}
&&R&0.8123&0.834&0.8517&0.8855&\textbf{0}&\textbf{0}&\textbf{0}&\textbf{0}\\
\cline{3-11}
&&OSR&\textbf{0.8176}&\textbf{0.8426}&\textbf{0.8708}&\textbf{0.9183}&\textbf{0}&\textbf{0}&\textbf{0}&\textbf{0}\\
\cline{3-11}
&&\cite{TCMK23}&0.8148&0.8204&0.807&0.6745&\textbf{0}&0.0445&0.0638&0.0848\\
\cline{3-11}
&&\cite{nature_feature}&0.8041&0.8154&0.8313&0.8593&\textbf{0}&\textbf{0}&\textbf{0}&\textbf{0}\\
\hhline{|~||==========}
&\multirow{5}{1.5cm}{MobileNet V2}&B&0.8066&0.8054&0.8045&0.805&0.9&0.816&0.812&0.8092\\
\cline{3-11}
&&R&\textbf{0.8080}&0.8233&0.8418&0.8795&\textbf{0}&\textbf{0}&\textbf{0}&\textbf{0}\\
\cline{3-11}
&&OSR&0.8070&\textbf{0.834}&\textbf{0.8603}&\textbf{0.9068}&\textbf{0}&\textbf{0}&\textbf{0}&\textbf{0}\\
\cline{3-11}
&&\cite{TCMK23}&0.8063&0.8018&0.7652&0.5928&\textbf{0}&0.065&0.0635&0.0435\\
\cline{3-11}
&&\cite{nature_feature}&0.7898&0.8074&0.8115&0.7988&\textbf{0}&\textbf{0}&\textbf{0}&\textbf{0}\\
\hline\hline
\multirow{10}{1cm}{VGG Face-100}&\multirow{5}{*}{ResNet18}&B&0.9415&0.9451&0.9462&0.9423&0.9882&0.9280&0.9349&0.9419\\
\cline{3-11}
&&R&0.9417&0.9429&0.9427&0.9413&\textbf{0}&\textbf{0}&\textbf{0}&\textbf{0}\\
\cline{3-11}
&&OSR&\textbf{0.9419}&\textbf{0.9530}&\textbf{0.9602}&\textbf{0.9629}&\textbf{0}&\textbf{0}&\textbf{0}&\textbf{0}\\
\cline{3-11}
&&\cite{TCMK23}&0.9380&0.9273&0.9215&0.9049&\textbf{0}&\textbf{0}&\textbf{0}&\textbf{0}\\
\cline{3-11}
&&\cite{nature_feature}&0.9341&0.9399&0.9261&0.6413&\textbf{0}&\textbf{0}&\textbf{0}&\textbf{0}\\
\hhline{|~||==========}
&\multirow{5}{1.5cm}{Vision Transformer B16}&B&0.9524&0.9547&0.9554&0.9520&0.9882&0.9444&0.9485&0.9535\\
\cline{3-11}
&&R&0.9482&0.9511&0.9571&0.9481&\textbf{0}&\textbf{0}&\textbf{0}&\textbf{0}\\
\cline{3-11}
&&OSR&\textbf{0.9527}&\textbf{0.9605}&\textbf{0.9652}&\textbf{0.9665}&\textbf{0}&\textbf{0}&\textbf{0}&\textbf{0}\\
\cline{3-11}
&&\cite{TCMK23}&0.9460&0.8935&0.8898&0.8008&1&0.7995&0.7619&0.4885\\
\cline{3-11}
&&\cite{nature_feature}&0.9514&0.0927&0.0435&0.0213&\textbf{0}&\textbf{0}&\textbf{0}&\textbf{0}\\
\hline
\end{tabular}
\end{scriptsize}
\centering
\end{table}

\begin{table}[t]
\centering
\caption{Comparison of coverage metric on CIFAR10, CIFAR100, and VGGFace-100 between base model (B), Retrained, (R), OSR (ours), UNSIR (\cite{TCMK23}), and NHLE-CWM (\cite{nature_feature}). Bolded entries show the best performance in each setting}
\label{tab:CM}
\begin{scriptsize}
\begin{tabular}{|p{0.1\linewidth}||p{0.12\linewidth}||p{0.09\linewidth}||p{0.07\linewidth}|p{0.07\linewidth}|p{0.07\linewidth}|p{0.07\linewidth}||p{0.07\linewidth}|p{0.07\linewidth}|p{0.07\linewidth}|p{0.07\linewidth}|}
\hline
\multirow{2}{*}{\textbf{Dataset}}&\multirow{2}{*}{\textbf{Model}}& \multirow{2}{*}{\textbf{Method}}&\multicolumn{4}{c||}{\textbf{Forget Classes}($\overline{Cov}_{\mathcal{R}}$) $\downarrow$}&\multicolumn{4}{c|}{\textbf{Forget Classes} ($\overline{Cov}_{\mathcal{F}}$) $\uparrow$}\\
\cline{4-11} 
&&&$\mathcal{F}_1$&$\mathcal{F}_2$&$\mathcal{F}_3$&$\mathcal{F}_4$&$\mathcal{F}_1$&$\mathcal{F}_2$&$\mathcal{F}_3$&$\mathcal{F}_4$\\
\hline
\hline
\multirow{10}{1cm}{CIFAR-10}&\multirow{5}{*}{ResNet18}&B&0.5222&0.5125&0.55&0.4667&0.6&0.6&0.5&0.5571\\
\cline{3-11}
&&R&0.5432&0.6094&0.6944&0.5556&$\boldsymbol{\infty}$&$\boldsymbol{\infty}$&$\boldsymbol{\infty}$&$\boldsymbol{\infty}$\\
\cline{3-11}
&&OSR&\textbf{0.4321}&\textbf{0.375}&\textbf{0.4444}&\textbf{0.4444}&$\boldsymbol{\infty}$&$\boldsymbol{\infty}$&$\boldsymbol{\infty}$&$\boldsymbol{\infty}$\\
\cline{3-11}
&&\cite{TCMK23}&0.5889&0.7&0.7167&0.8333&0.9&0.85&0.85&0.8429\\
\cline{3-11}
&&\cite{nature_feature}&0.5556&0.725&0.7&0.8667&0.7&0.65&0.6&0.3\\
\hhline{|~||==========}
&\multirow{5}{*}{AllCNN}&B&0.6333&0.625&0.6333&0.6333&0.7&0.7&0.65&0.6429\\
\cline{3-11}
&&R&0.6049&0.5625&0.6667&0.6667&$\boldsymbol{\infty}$&$\boldsymbol{\infty}$&$\boldsymbol{\infty}$&$\boldsymbol{\infty}$\\
\cline{3-11}
&&OSR &\textbf{0.4198}&\textbf{0.3906}&\textbf{0.4444}&\textbf{0.4444}&$\boldsymbol{\infty}$&$\boldsymbol{\infty}$&$\boldsymbol{\infty}$&$\boldsymbol{\infty}$\\
\cline{3-11}
&&\cite{TCMK23}&0.7778&0.675&0.85&0.8667&0.9&0.9&0.875&0.8857\\
\cline{3-11}
&&\cite{nature_feature}&0.6222&0.625&0.8167&0.9&0.7&0.65&0.575&0.3\\
\hline\hline
\multirow{10}{1cm}{CIFAR-100}&\multirow{5}{*}{ResNet18}&B&0.2016&0.2059&0.1827&0.1788&0.04&0.1765&0.2260&0.2142\\
\cline{3-11}
&&R&0.1923&0.1936&0.1675&0.1875&$\boldsymbol{\infty}$&$\boldsymbol{\infty}$&$\boldsymbol{\infty}$&$\boldsymbol{\infty}$\\
\cline{3-11}
&&OSR&\textbf{0.1749}&\textbf{0.0506}&\textbf{0.0436}&\textbf{0.0388}&$\boldsymbol{\infty}$&$\boldsymbol{\infty}$&$\boldsymbol{\infty}$&$\boldsymbol{\infty}$\\
\cline{3-11}
&&\cite{TCMK23}&0.2118&0.2403&0.2495&0.3830&0.99&0.919&0.9003&0.8458\\
\cline{3-11}
&&\cite{nature_feature}&0.2173&0.1960&0.1423&0.1155&0.51&0.4555&0.3688&0.2670\\
\hhline{|~||==========}
&\multirow{5}{1.5cm}{MobileNet V2}&B&0.1959&0.2020&0.1763&0.1808&0.04&0.1635&0.2213&0.2033\\
\cline{3-11}
&&R&\textbf{0.1670}&0.2069&0.1781&0.1938&$\boldsymbol{\infty}$&$\boldsymbol{\infty}$&$\boldsymbol{\infty}$&$\boldsymbol{\infty}$\\
\cline{3-11}
&&OSR&0.1750&\textbf{0.0516}&\textbf{0.0433}&\textbf{0.0444}&$\boldsymbol{\infty}$&$\boldsymbol{\infty}$&$\boldsymbol{\infty}$&$\boldsymbol{\infty}$\\
\cline{3-11}
&&\cite{TCMK23}&0.2036&0.2504&0.2732&0.4033&0.9900&0.7990&0.8698&0.871\\
\cline{3-11}
&&\cite{nature_feature}&0.217&0.1861&0.1737&0.2803&0.52&0.451&0.3618&0.2662\\
\hline\hline
\multirow{10}{1cm}{VGG Face-100}&\multirow{5}{*}{ResNet18}&B&0.0741&0.0733&0.0763&0.0758&0.01&0.0745&0.0693&0.072\\
\cline{3-11}
&&R&0.0915&0.0883&0.1017&0.1456&$\boldsymbol{\infty}$&$\boldsymbol{\infty}$&$\boldsymbol{\infty}$&$\boldsymbol{\infty}$\\
\cline{3-11}
&&OSR&\textbf{0.0691}&\textbf{0.0339}&\textbf{0.0319}&\textbf{0.0375}&$\boldsymbol{\infty}$&$\boldsymbol{\infty}$&$\boldsymbol{\infty}$&$\boldsymbol{\infty}$\\
\cline{3-11}
&&\cite{TCMK23}&0.0817&0.1028&0.1247&0.1515&0.98&0.8295&0.9063&0.9213\\
\cline{3-11}
&&\cite{nature_feature}&0.0855&0.0715&0.0675&0.2890&0.52&0.4510&0.3638&0.2767\\
\hhline{|~||==========}
&\multirow{5}{1.5cm}{Vision Transformer B16}&B&0.0727&0.0813&0.0755&0.0693&0.01&0.0355&0.067&0.074\\
\cline{3-11}
&&R&0.0725&0.0684&0.1&0.1356&$\boldsymbol{\infty}$&$\boldsymbol{\infty}$&$\boldsymbol{\infty}$&$\boldsymbol{\infty}$\\
\cline{3-11}
&&OSR&\textbf{0.0659}&\textbf{0.0363}&\textbf{0.0347}&\textbf{0.0419}&$\boldsymbol{\infty}$&$\boldsymbol{\infty}$&$\boldsymbol{\infty}$&$\boldsymbol{\infty}$\\
\cline{3-11}
&&\cite{TCMK23}&0.0811&0.1333&0.1483&0.1908&0&0.1045&0.1813&0.3382\\
\cline{3-11}
&&\cite{nature_feature}&0.073&0.5169&0.5505&0.4138&0.5&0.5375&0.4655&0.3603\\
\hline
\end{tabular}
\end{scriptsize}
\centering
\end{table}

\begin{table}[t]
\centering
\caption{Comparison of KL Divergence on CIFAR10, CIFAR100, and VGGFace-100 between retrained model (R) against OSR (ours), UNSIR (\cite{TCMK23}), and NHLE-CWM (\cite{nature_feature}). Bolded entries show the best performance in each setting. For each metric, the arrow indicates whether better solutions have larger ($\uparrow$) or smaller ($\downarrow$) value.}
\label{tab:KLRM}
\begin{scriptsize}
\begin{tabular}{|p{0.1\linewidth}||p{0.12\linewidth}||p{0.09\linewidth}||p{0.07\linewidth}|p{0.07\linewidth}|p{0.07\linewidth}|p{0.07\linewidth}||p{0.07\linewidth}|p{0.07\linewidth}|p{0.07\linewidth}|p{0.07\linewidth}|}
\hline
\multirow{2}{*}{\textbf{Dataset}}&\multirow{2}{*}{\textbf{Model}}& \multirow{2}{*}{\textbf{Method}}&\multicolumn{4}{c||}{\textbf{Forget Classes}($KL^R_{\mathcal{R}}$) $\downarrow$}&\multicolumn{4}{c|}{\textbf{Forget Classes} ($KL^R_{\mathcal{F}}$) $\downarrow$}\\
\cline{4-11} 
&&&$\mathcal{F}_1$&$\mathcal{F}_2$&$\mathcal{F}_3$&$\mathcal{F}_4$&$\mathcal{F}_1$&$\mathcal{F}_2$&$\mathcal{F}_3$&$\mathcal{F}_4$\\
\hline
\hline
\multirow{6}{1cm}{CIFAR-10}&\multirow{3}{*}{ResNet18}&OSR&\textbf{0.1784}&\textbf{0.1729}&\textbf{0.2316}&\textbf{0.0888}&\textbf{1.082}&\textbf{1.2932}&\textbf{1.6925}&2.1947\\
\cline{3-11}
&&\cite{TCMK23}&0.5683&0.6290&1.281&0.3616&1.6559&1.9389&3.4369&\textbf{0.8137}\\
\cline{3-11}
&&\cite{nature_feature}&0.4157&0.444&1.386&0.9566&2.1313&2.7903&4.8446&8.0864\\
\hhline{|~||==========}
&\multirow{3}{*}{AllCNN}&OSR &\textbf{0.1102}&\textbf{0.1020}&\textbf{0.1324}&\textbf{0.0851}&\textbf{0.6176}&\textbf{0.5881}&\textbf{0.6663}&1.2840\\
\cline{3-11}
&&\cite{TCMK23}&0.4003&0.4504&0.6379&0.4836&0.8151&1.0849&1.3144&\textbf{1.2034}\\
\cline{3-11}
&&\cite{nature_feature}&0.2450&0.1633&1.8146&1.2182&1.2961&1.0497&3.4006&5.4858\\
\hline\hline
\multirow{6}{1cm}{CIFAR-100}&\multirow{3}{*}{ResNet18}&OSR&\textbf{0.1733}&\textbf{0.1961}&\textbf{0.2217}&\textbf{0.2261}&\textbf{0.3272}&\textbf{0.6041}&\textbf{0.9072}&\textbf{1.1154}\\
\cline{3-11}
&&\cite{TCMK23}&0.1818&0.2&0.2585&0.5685&0.3926&2.4210&5.1018&8.9134\\
\cline{3-11}
&&\cite{nature_feature}&0.2445&0.2349&0.2693&0.3231&0.6308&0.6554&0.9578&1.4806\\
\hhline{|~||==========}
&\multirow{3}{1.5cm}{MobileNet V2}&OSR&0.1987&\textbf{0.2237}&\textbf{0.2540}&\textbf{0.2704}&\textbf{0.2853}&\textbf{0.7122}&\textbf{1.0202}&\textbf{1.2177}\\
\cline{3-11}
&&\cite{TCMK23}&\textbf{0.1981}&0.26&0.4296&1.2407&0.3177&2.601&4.8616&8.4283\\
\cline{3-11}
&&\cite{nature_feature}&0.3014&0.2712&0.3911&0.938&0.4268&0.7965&1.3189&2.7282\\
\hline\hline
\multirow{6}{1cm}{VGG Face-100}&\multirow{3}{*}{ResNet18}&OSR&\textbf{0.1631}&\textbf{0.1405}&\textbf{0.1545}&\textbf{0.1580}&\textbf{0.8174}&\textbf{0.8886}&\textbf{0.9698}&\textbf{1.1304}\\
\cline{3-11}
&&\cite{TCMK23}&0.1946&0.1881&0.2076&0.2337&0.9632&5.13&6.4925&8.2943\\
\cline{3-11}
&&\cite{nature_feature}&0.2060&0.2277&0.5316&3.7939&0.9889&1.2484&1.7372&3.8979\\
\hhline{|~||==========}
&\multirow{3}{1.5cm}{Vision Transformer B16}&OSR&\textbf{0.0996}&\textbf{0.0907}&\textbf{0.0943}&\textbf{0.1191}&1.1588&\textbf{0.9416}&\textbf{1.0542}&\textbf{1.3507}\\
\cline{3-11}
&&\cite{TCMK23}&0.1244&0.7858&0.8738&2.2414&9.0109&5.5759&5.6367&4.0039\\
\cline{3-11}
&&\cite{nature_feature}&0.1002&7.3222&7.89&8.7059&\textbf{1.0712}&4.6728&4.5615&6.4601\\
\hline
\end{tabular}
\end{scriptsize}
\centering
\end{table}

\begin{table}[h]
\centering
\caption{Comparison of KL Divergence on CIFAR10, CIFAR100, and VGGFace-100 between base model (B) against OSR (ours), UNSIR (\cite{TCMK23}), and NHLE-CWM (\cite{nature_feature}). Bolded entries show the best performance in each setting. For each metric, the arrow indicates whether better solutions have larger ($\uparrow$) or smaller ($\downarrow$) value.}
\label{tab:KLBM}
\begin{scriptsize}
\begin{tabular}{|p{0.1\linewidth}||p{0.12\linewidth}||p{0.09\linewidth}||p{0.07\linewidth}|p{0.07\linewidth}|p{0.07\linewidth}|p{0.07\linewidth}||p{0.07\linewidth}|p{0.07\linewidth}|p{0.07\linewidth}|p{0.07\linewidth}|}
\hline
\multirow{2}{*}{\textbf{Dataset}}&\multirow{2}{*}{\textbf{Model}}& \multirow{2}{*}{\textbf{Method}}&\multicolumn{4}{c||}{\textbf{Forget Classes}($KL^B_{\mathcal{R}}$) $\downarrow$}&\multicolumn{4}{c|}{\textbf{Forget Classes} ($KL^B_{\mathcal{F}}$) $\uparrow$}\\
\cline{4-11} 
&&&$\mathcal{F}_1$&$\mathcal{F}_2$&$\mathcal{F}_3$&$\mathcal{F}_4$&$\mathcal{F}_1$&$\mathcal{F}_2$&$\mathcal{F}_3$&$\mathcal{F}_4$\\
\hline
\hline
\multirow{6}{1cm}{CIFAR-10}&\multirow{3}{*}{ResNet18}&OSR&\textbf{0.0030}&\textbf{0.0087}&\textbf{0.0154}&\textbf{0.0077}&8.9727&9.3584&10.3125&11.5819\\
\cline{3-11}
&&\cite{TCMK23}&0.4629&0.6060&1.1698&0.4203&\textbf{9.9941}&\textbf{10.588}&\textbf{11.7677}&\textbf{12.1166}\\
\cline{3-11}
&&\cite{nature_feature}&0.1955&0.2739&1.0356&0.6971&9.4887&10.1972&11.6639&11.4442\\
\hhline{|~||==========}
&\multirow{3}{*}{AllCNN}&OSR &\textbf{0.0072}&\textbf{0.0132}&\textbf{0.0195}&\textbf{0.0142}&5.1187&6.4038&7.8772&9.2597\\
\cline{3-11}
&&\cite{TCMK23}&0.3336&0.4902&0.6867&0.7073&5.4451&\textbf{6.9287}&8.5991&\textbf{9.6862}\\
\cline{3-11}
&&\cite{nature_feature}&0.1061&0.0749&1.7802&1.3557&\textbf{5.4616}&6.5652&\textbf{9.8556}&9.4157\\
\hline\hline
\multirow{6}{1cm}{CIFAR-100}&\multirow{3}{*}{ResNet18}&OSR&\textbf{0.0004}&\textbf{0.0341}&\textbf{0.064}&\textbf{0.0880}&4.1952&3.938&4.7698&5.6617\\
\cline{3-11}
&&\cite{TCMK23}&0.0268&0.0412&0.089&0.4346&4.2453&\textbf{4.5003}&\textbf{6.1877}&\textbf{8.0276}\\
\cline{3-11}
&&\cite{nature_feature}&0.0652&0.0662&0.0828&0.1235&\textbf{4.4348}&3.9725&4.6271&5.3763\\
\hhline{|~||==========}
&\multirow{3}{1.5cm}{MobileNet V2}&OSR&\textbf{0.0005}&\textbf{0.0357}&\textbf{0.0688}&\textbf{0.0986}&4.6331&3.9890&4.7484&5.5058\\
\cline{3-11}
&&\cite{TCMK23}&0.0213&0.0832&0.2314&1.0471&4.646&\textbf{4.1023}&\textbf{5.7880}&\textbf{7.978}\\
\cline{3-11}
&&\cite{nature_feature}&0.0793&0.0767&0.1459&0.6394&\textbf{4.8304}&4.0432&4.6947&5.8235\\
\hline\hline
\multirow{6}{1cm}{VGG Face-100}&\multirow{3}{*}{ResNet18}&OSR&\textbf{0.0001}&\textbf{0.0058}&\textbf{0.011}&\textbf{0.0156}&4.3195&4.2046&4.8145&5.7763\\
\cline{3-11}
&&\cite{TCMK23}&0.0316&0.0459&0.0565&0.0821&\textbf{4.5182}&\textbf{7.9142}&\textbf{9.6922}&\textbf{10.9717}\\
\cline{3-11}
&&\cite{nature_feature}&0.0340&0.0771&0.3467&3.9801&4.4576&4.4413&5.2942&7.1377\\
\hhline{|~||==========}
&\multirow{3}{1.5cm}{Vision Transformer B16}&OSR&\textbf{<0.0001}&\textbf{0.0029}&\textbf{0.0056}&\textbf{0.0079}&5.2295&4.6097&5.2415&6.0458\\
\cline{3-11}
&&\cite{TCMK23}&0.0241&0.7553&0.8393&2.5364&0.0187&1.2922&1.5609&4.2886\\
\cline{3-11}
&&\cite{nature_feature}&0.0013&7.3263&7.9734&8.9262&\textbf{5.2324}&\textbf{7.4741}&\textbf{7.6715}&\textbf{9.6077}\\
\hline
\end{tabular}
\end{scriptsize}
\centering
\end{table}

Summing over all steps, the overall complexity of our proposed method is \(\mathcal{C}=(N_i+1)\cdot\mathcal{C}(f)+(N_i+4)\cdot\mathcal{C}_{\mathtt{Div}}+O\left(n^3+N_in^2\right).\) Since $N > N_i \gg n$ in our setting, this simplifies to\(
\mathcal{C}= (N_i+1)\cdot \mathcal{C}(f) + (N_i+4)\cdot\mathcal{C}_{\mathtt{Div}}+\theta(N_in^2).    
\) Based on this approximation of the overall complexity, asymptotically we require the same number of inference computation (referring to the computation of $f$) as to the division operations. Similarly, we asymptotically require $O(n^2)$ times more multiplication operations than the call to $f.$ However, in practice, we can assume that one call to the the model computation is much more complex than requiring just $1$ division operation or $n^2$ multiplications. Hence, we can expect that the dominant cost of our proposed solution is $O(N_i)C(f).$

\section{Experiments and Analyses}
In this section, we consider the performance of our proposed solution along with its comparison to existing baseline methods.

\subsection{Dataset}
In our experiment, we consider the implementation of our proposed solution in the following datasets, image classification models, and sets of removed-labels.
\begin{enumerate}
    \item CIFAR-10 \cite{krizhevsky2009cifar} that is trained from scratch using the image classification models ResNet18 \cite{he2016resnet} and AllCNN \cite{springenberg2014allcnn}. In our experiment, among the 10 labels $\{0,\cdots, 9\},$ we consider four cases for the sets of the labels to be removed, namely $\{0\}, \{0,4\},$ $\{0,4,7,9\},$ and $\{0,2,3,4,6,7,9\}.$ We call these forget sets $\mathcal{F}_1,\cdots, \mathcal{F}_4$ respectively.
    \item CIFAR-100 \cite{krizhevsky2009cifar} fine-tuned from the image classification models ResNet18 and MobileNetV2 \cite{sandler2018mobilenetv2} which are pre-trained using the ImageNet-1K \cite{deng2009imagenet}. For CIFAR-100 dataset with labels denoted by $\{0,\cdots, 99\}$, we consider four cases for the sets of the labels to be removed, $\{0,1,\cdots, i-1\}$ for $i\in \{1,20,40,60\}.$ We call these forget sets $\mathcal{F}_1,\cdots, \mathcal{F}_4$ respectively.
    \item VGGFace-100\footnote{A 100-identity subset of VGGFace2~\cite{cao2018vggface2}, constructed by randomly sampling 100 identities and their associated images from the full dataset.} fine-tuned from the image classification models ResNet18 and Vision Transformer B16 \cite{dosovitskiy2020vit} which are pre-trained with ImageNet-1K. Similar to the sets of removed labels in CIFAR-100, the $100$ labels are denoted by $\{0,\cdots, 99\}$ and we consider four different sets of the removed labels $\{0,\cdots, i-1\}$ for $i\in \{1,20,40,60\}.$ We call these forget sets $\mathcal{F}_1,\cdots, \mathcal{F}_4$ respectively.
\end{enumerate}

\subsection{Baseline Methods}
In our experiments, we compare OSR with state-of-the-art machine unlearning methods, namely UNSIR \cite{TCMK23} and NHLE-CWM \cite{nature_feature}. We consider the desired target behavior to be the model obtained from the approach of retraining the model with the whole dataset without those with their true label belonging to the set of removed labels. For UNSIR and NHLE-CWM, we implement their proposed algorithms following the parameters stated in their respective works \cite{TCMK23,nature_feature} prioritizing our implementation to reproduce their best performance with the following remark.
\begin{itemize}
    \item For our implementation of UNSIR:
    \begin{itemize}
        \item Instead of the use of Adam \cite{kingma2014adam} as the optimizer, we consider the use of the more powerful SGD \cite{sutskever2013importance} as the optimizer to provide a fair comparison with our approach that uses SGD optimizer. Without such alteration, it causes the performance of UNSIR to be significantly lower.
        \item For the implementation on VGGFace-100, we reduce the noise parameter from $0.1$ to $0.001$ to account for the 49 times larger L2 term for 224$\times$ 224 inputs compared to that for 32$\times$32 inputs.
        \item For the implementation on CIFAR-100, we increase the sample size used for each batch from $50$ to $100$ while for the implementation on VGGFace-100, the batch size is increased from $100$ to $500.$ This alteration is needed to ensure that the repair step of UNSIR covers sufficient amount of data to perform in an acceptable level.
        \end{itemize}
        \item For our implementation of NHLE-CWM:
        \begin{itemize}
            \item We set the learning rates to the following $0.001$ for CIFAR-10, $0.0001$ for CIFAR-100, and $0.0001$ on VGGFace100. This is different from the implementation described in \cite{nature_feature}, which implies the use of $0.01$ for all implementations. This change is done because using the original learning rate causes runaway gradient problem in our experiment.
            \item When considering removal of more than $20$ labels, to avoid the significant increase in the loss value, we run NHLE-CWM for $3$ epochs instead of $10$ epochs. 
        \end{itemize}
\end{itemize}

Our experiments are conducted on a virtual machine equipped with two NVIDIA A100 (80 GB) GPUs, AMPD EPYC 7763 64-core, and 64GB RAM running Ubuntu 22.04. 

\subsection{Evaluation Metrics}
We briefly discuss the evaluation metrics defined in the works on machine unlearning  \cite{TCMK23,nature_feature} and multilabel learning \cite{GV15} that will be used as our evaluation metrics.

\begin{enumerate}
    \item Accuracy. Based on the output after the application of the solutions, we may take the label with the highest confidence value to be the prediction of the classifier. Hence, for each dataset $\mathcal{D}$ and each possible label value $i$, we measure the proportion of $\mathcal{D}_i$ that is correctly predicted to have label $i.$ We are mainly interested in finding the average of such proportions for the labels retained, denoted by $\mathcal{A}_{R},$ as well as the average of such proportions for the removed labels, denoted by $\mathcal{A}_F$. It is clear that the better solution should have higher $\mathcal{A}_R$ while $\mathcal{A}_F$ should be $0.$
    \item Coverage. Coverage is a metric that is similar to the accuracy metric. However, instead of focusing only on the label with the highest confidence, we are more interested to measure \emph{the position} of the true label when the labels are sorted based on its confidence value in the output of the approach. It is clear that if the average of such position is small, then even when the label with highest confidence is not the true label, the true label is still predicted with high probability. More specifically, for each $x\in \mathcal{D}$ with true label $i$ which is not removed, for an approach $F,$ we define $r_{F,x,i}\in \mathbb{Z}_{>0}$ such that $F(x)$ has its $i$-th  value to be the $r_{F,x,i}$-th largest value among the entries of $F(x).$ We further define $r_{F,i}=\max_{x\in \mathcal{D}_i}r_{F,x,i},$ the maximum value of $r_{F,x,i}$ among all $x\in \mathcal{D}_i.$ Lastly, the coverage of $F,$ denoted by $r_{F}$ is defined to be the average of $r_{F,i}$ for all $i\in \mathcal{L}_r.$ To normalize the value so that a perfect $F$ gives $Cov_{\mathcal{R}}(F)=0,$ we define $Cov_{\mathcal{R}}(F)$ to be $r_F-1.$ That is:
    \(Cov_{\mathcal{R}}(F)=\frac{1}{|\mathcal{L}_r|}\sum_{i\in \mathcal{L}_r}\max\{r_{F,x,i}:x\in \mathcal{D}_i\}-1.\)
    We extend the definition of coverage to also consider the removed labels. More specifically, we define $Cov_{\mathcal{F}}(F)$ similarly where instead of considering the non-removed labels $\mathcal{L}_r,$ we consider the removed labels $\mathcal{L}_f.$ Here for solutions that simply remove the forgotten labels, we simply set it as $\infty.$ 
    \item KL divergence \cite{kullback1951kl} of confidence vectors. Beyond the top label, we quantify the similarity of full output vectors to the retraining baseline; smaller average divergences indicate closer approximation. We report two metrics: (a) retained-label inputs and (b) forgotten-label inputs. A good label-removal method should preserve information about non-removed labels (low divergence from the base model) while removing information about removed labels (high divergence when the input is from a removed label). 
    \item Processing Time. We are interested in measuring the time required by different approaches in completing the label removal process. It is clear that the smaller the processing time, the better from the perspective of its feasibility to be applied in a real-life application. Here, for different approaches, we measure the time required for each step to better understand the bottleneck of different approaches.
\end{enumerate}

We note that we omit two metrics considered in \cite{TCMK23,nature_feature}, namely the retraining time and weight distance. We recall that we consider the problem of label removal from the perspective of the output distribution instead of the perspective of the model producing such output. So instead of modifying the trained model to approximate the retraining approach, our approach serves as a post-hoc procedure to process the output, which can be seen as a filter to be applied to existing models. This leads to the following observations. Firstly, since our approach applies a filter to the output of the original model, to recover the original model, we can simply remove the filter. Hence, retraining time is trivial. Similarly, the weights for the parameters of the model never change, hence the distance is $0.$ We note that because of this, our approach does not suffer from the risk of catastrophic forgetting that may happen in other approaches, especially when the number of removed labels increases.

\subsection{Experimental Results}
In this section, we discuss some of the measurement results from our experiments. 
\subsubsection{Accuracy}
As has been previously discussed, we first consider the comparison of both retain accuracy $\mathcal{A}_{\mathcal{R}}$ and forget accuracy $\mathcal{A}_{\mathcal{F}}$ in different datasets and models. We also consider four different forget set with increasing size for each dataset where the forgotten set is sampled randomly with the requirement that $\mathcal{F}_1\subseteq\mathcal{F}_2\subseteq\mathcal{F}_3\subseteq\mathcal{F}_4.$ As can be observed in Table \ref{tab:AM}, OSR outperforms the state-of-the-art solutions UNSIR \cite{TCMK23} and NHLE-CWM \cite{nature_feature} in both $\mathcal{A}_{\mathcal{R}}$ and $\mathcal{A}_{\mathcal{F}},$ which shows that after applying our solutions, the resulting output still preserves very high accuracy in retained labels and it has 0 probability to still predict the forgotten labels. We note that in most cases, our accuracy in $\mathcal{A}_{\mathcal{R}}$ is higher than both the base model (before forgetting) and the retrained model (where a fresh model is trained just using the training data belonging to the retained labels). This can be argued in the following manner. Compared to the baseline model, since some labels are removed, data from a remaining label that is close to the removed label may no longer be misclassified as the label that is being removed. Because of this, we can expect a better separation between remaining labels. This can also be observed in the fact that most retrained models have better average accuracy than the base model. Secondly, OSR outperforms the accuracy of the retrained model due to the fact that the information that can be learned from the removed label will still help our solution in refining the prediction of the remaining labels better, which can be observed from the higher $\mathcal{A}_{\mathcal{R}}$ of OSR compared to that of the retrained model. 

\begin{table}[t]
\centering
\caption{Comparison of Processing Time (in seconds) on VGGFace-100 between retrained model (R), OSR (ours), UNSIR (\cite{TCMK23}), and NHLE-CWM (\cite{nature_feature}). For Retraining method, Ave: Average per Epoch, Tot (n): Total for $n$ Epochs. For OSR, Pro: Projection, Red: Redistribution, Tot: Total. For UNSIR, NG: Noise Generation, RC: Retain Collection, Im: Impair, Re: Repair, Tot: Total. For NHLE-CWM, FT: Finetune, CWM: Class Weight Masking, Tot: Total
}
\begin{scriptsize}
\label{tab:TimeVGGFace100}
\begin{tabular}
{|p{0.1\linewidth}||p{0.1\linewidth}|p{0.15\linewidth}||p{0.15\linewidth}|p{0.15\linewidth}|p{0.15\linewidth}|p{0.12\linewidth}|}
\hline
\multirow{2}{*}{\textbf{Model}}& \multirow{2}{*}{\textbf{Method}}&\multirow{2}{*}{\textbf{Component}}&\multicolumn{4}{c|}{\textbf{Forget Classes}(Time) $\downarrow$}\\
\cline{4-7} 
&&&$\mathcal{F}_1$&$\mathcal{F}_2$&$\mathcal{F}_3$&$\mathcal{F}_4$\\
\hline
\hline
\multirow{13}{*}{ResNet18}&\multirow{2}{*}{R}&Ave&130.4148&109.0052&84.2198&57.9094\\
\cline{3-7}
&&\cellcolor{gray!50} Tot(5) &\cellcolor{gray!50} 652.074 & \cellcolor{gray!50} 545.026 & \cellcolor{gray!50} 421.099 & \cellcolor{gray!50} 289.547\\
\hhline{|~||======}
&\multirow{3}{*}{OSR}&Pro&0.145&0.306&0.494&0.594\\
\cline{3-7}
&&Red&0.022&0.021&0.039&0.018\\
\cline{3-7}
&&\cellcolor{gray!50}Total&\cellcolor{gray!50}\textbf{0.167}&\cellcolor{gray!50}\textbf{0.327}&\cellcolor{gray!50}\textbf{0.533}&\cellcolor{gray!50}\textbf{0.612}\\
\hhline{|~||======}
&\multirow{5}{*}{\cite{TCMK23}}&NG&52.435&1044.551&2088.334&3132.282\\
\cline{3-7}
&&RC&19.409&18.87&20.14&20.342\\
\cline{3-7}
&&Im&651.557&1753.912&1977.476&2588.275\\
\cline{3-7}
&&Re&563.656&271.413&175.297&176.551\\
\cline{3-7}
&&\cellcolor{gray!50}Tot&\cellcolor{gray!50}1287.057&\cellcolor{gray!50}3088.746&\cellcolor{gray!50}4261.247&\cellcolor{gray!50}5917.45\\
\hhline{|~||======}
&\multirow{3}{*}{\cite{nature_feature}}&FT&3.662&16.283&30.635&44.172\\
\cline{3-7}
&&CWM&0.0&0.002&0.001&0.002\\
\cline{3-7}
&&\cellcolor{gray!50}Tot&\cellcolor{gray!50}3.662&\cellcolor{gray!50}16.285&\cellcolor{gray!50}30.636&\cellcolor{gray!50}44.174\\
\hhline{|=||======}
\multirow{13}{1.5cm}{Vision Transformer B16}&\multirow{2}{*}{R}&Ave&25.5474&22.0204&17.8888&12.7862\\
\cline{3-7}
&&\cellcolor{gray!50} Tot(5) &\cellcolor{gray!50} 127.737 & \cellcolor{gray!50} 110.102 & \cellcolor{gray!50} 89.444 & \cellcolor{gray!50} 63.931\\
\hhline{|~||======}
&\multirow{3}{*}{OSR}&Pro&0.137&0.2&0.229&0.242\\
\cline{3-7}
&&Red&0.021&0.02&0.023&0.021\\
\cline{3-7}
&&\cellcolor{gray!50}Total&\cellcolor{gray!50}\textbf{0.158}&\cellcolor{gray!50}\textbf{0.22}&\cellcolor{gray!50}\textbf{0.252}&\cellcolor{gray!50}\textbf{0.263}\\
\hhline{|~||======}
&\multirow{5}{*}{\cite{TCMK23}}&NG&3.807&61.821&124.053&185.884\\
\cline{3-7}
&&RC&18.747&20.786&19.795&19.733\\
\cline{3-7}
&&Im&206.522&969.596&1005.548&1216.515\\
\cline{3-7}
&&Re&42.992&21.597&80.029&9.531\\
\cline{3-7}
&&\cellcolor{gray!50}Tot&\cellcolor{gray!50}272.068&\cellcolor{gray!50}1073.8&\cellcolor{gray!50}1229.425&\cellcolor{gray!50}1431.663\\
\hhline{|~||======}
&\multirow{3}{*}{\cite{nature_feature}}&FT&3.226&12.571&23.646&34.595\\
\cline{3-7}
&&CWM&0.0&0.001&0.001&0.002\\
\cline{3-7}
&&\cellcolor{gray!50}Tot&\cellcolor{gray!50}3.226&\cellcolor{gray!50}12.572&\cellcolor{gray!50}23.647&\cellcolor{gray!50}34.597\\
\hline
\end{tabular}
\end{scriptsize}
\centering
\end{table}

\begin{table}[t]
\centering
\caption{Comparison of Processing Time (in seconds) on CIFAR10 between retrained model (R), OSR (ours), UNSIR (\cite{TCMK23}), and NHLE-CWM (\cite{nature_feature}). For Retraining method, Ave: Average per Epoch, Tot (n): Total for $n$ Epochs. For OSR, Pro: Projection, Red: Redistribution, Tot: Total. For UNSIR, NG: Noise Generation, RC: Retain Collection, Im: Impair, Re: Repair, Tot: Total. For NHLE-CWM, FT: Finetune, CWM: Class Weight Masking, Tot: Total}
\label{tab:TimeCIFAR10}
\begin{scriptsize}
\begin{tabular}{|p{0.1\linewidth}||p{0.1\linewidth}|p{0.15\linewidth}||p{0.15\linewidth}|p{0.15\linewidth}|p{0.15\linewidth}|p{0.12\linewidth}|}
\hline
\multirow{2}{*}{\textbf{Model}}& \multirow{2}{*}{\textbf{Method}}&\multirow{2}{*}{\textbf{Component}}&\multicolumn{4}{c|}{\textbf{Forget Classes}(Time) $\downarrow$}\\
\cline{4-7} 
&&&$\mathcal{F}_1$&$\mathcal{F}_2$&$\mathcal{F}_3$&$\mathcal{F}_4$\\
\hline
\hline
\multirow{13}{*}{ResNet18}&\multirow{2}{*}{R}&Ave&7.9859&7.8332&7.3648&5.7423\\
\cline{3-7}
&&\cellcolor{gray!50} Tot(40) &\cellcolor{gray!50} 319.437 & \cellcolor{gray!50} 313.328 & \cellcolor{gray!50} 294.592 & \cellcolor{gray!50}5.7423\\
\hhline{|~||======}
&\multirow{3}{*}{OSR}&Pro&0.004&0.037&0.005&0.003\\
\cline{3-7}
&&Red&0.004&0.016&0.004&0.044\\
\cline{3-7}
&&\cellcolor{gray!50}Total&\cellcolor{gray!50}\textbf{0.008}&\cellcolor{gray!50}\textbf{0.053}&\cellcolor{gray!50}\textbf{0.009}&\cellcolor{gray!50}\textbf{0.007}\\
\hhline{|~||======}
&\multirow{5}{*}{\cite{TCMK23}}&NG&1.628&1.94&3.974&6.821\\
\cline{3-7}
&&RC&1.082&1.09&1.08&1.155\\
\cline{3-7}
&&Im&2.624&3.049&4.475&6.203\\
\cline{3-7}
&&Re&1.46&1.333&0.879&0.386\\
\cline{3-7}
&&\cellcolor{gray!50}Tot&\cellcolor{gray!50}6.794&\cellcolor{gray!50}7.412&\cellcolor{gray!50}10.408&\cellcolor{gray!50}14.565\\
\hhline{|~||======}
&\multirow{3}{*}{\cite{nature_feature}}&FT&0.809&0.458&0.531&0.685\\
\cline{3-7}
&&CWM&0.0&0.0&0.0&0.0\\
\cline{3-7}
&&\cellcolor{gray!50}Tot&\cellcolor{gray!50}0.809&\cellcolor{gray!50}0.458&\cellcolor{gray!50}0.531&\cellcolor{gray!50}0.685\\
\hhline{|=||======}
\multirow{13}{*}{AllCNN}&\multirow{2}{*}{R}&Ave&5.1727&5.6598&5.1892&4.4128\\
\cline{3-7}
&&\cellcolor{gray!50} Tot(40) &\cellcolor{gray!50} 206.909 & \cellcolor{gray!50} 226.391 & \cellcolor{gray!50} 207.566 & \cellcolor{gray!50} 176.511\\
\hhline{|~||======}
&\multirow{3}{*}{OSR}&Pro&0.004&0.03&0.003&0.003\\
\cline{3-7}
&&Red&0.032&0.004&0.004&0.044\\
\cline{3-7}
&&\cellcolor{gray!50}Total&\cellcolor{gray!50}\textbf{0.036}&\cellcolor{gray!50}\textbf{0.034}&\cellcolor{gray!50}\textbf{0.007}&\cellcolor{gray!50}\textbf{0.007}\\
\hhline{|~||======}
&\multirow{5}{*}{\cite{TCMK23}}&NG&0.951&0.996&1.989&3.377\\
\cline{3-7}
&&RC&1.104&1.112&1.097&1.135\\
\cline{3-7}
&&Im&1.614&1.505&2.43&3.697\\
\cline{3-7}
&&Re&0.807&0.832&0.464&0.229\\
\cline{3-7}
&&\cellcolor{gray!50}Tot&\cellcolor{gray!50}4.476&\cellcolor{gray!50}4.445&\cellcolor{gray!50}5.98&\cellcolor{gray!50}8.438\\
\hhline{|~||======}
&\multirow{3}{*}{\cite{nature_feature}}&FT&0.826&0.408&0.489&0.596\\
\cline{3-7}
&&CWM&0.0&0.0&0.0&0.0\\
\cline{3-7}
&&\cellcolor{gray!50}Tot&\cellcolor{gray!50}0.826&\cellcolor{gray!50}0.408&\cellcolor{gray!50}0.489&\cellcolor{gray!50}0.596\\
\hline
\end{tabular}
\end{scriptsize}
\centering
\end{table}

\begin{table}[t]
\centering
\caption{Comparison of Processing Time (in seconds) on CIFAR100 between retrained model (R), OSR (ours), UNSIR (\cite{TCMK23}), and NHLE-CWM (\cite{nature_feature}). 
 Bolded entries show the best performance in each setting. For each metric, the arrow indicates whether better solutions have larger ($\uparrow$) or smaller ($\downarrow$) value. For Retraining method, Ave: Average per Epoch, Tot (n): Total for $n$ Epochs. For OSR, Pro: Projection, Red: Redistribution, Tot: Total. For UNSIR, NG: Noise Generation, RC: Retain Collection, Im: Impair, Re: Repair, Tot: Total. For NHLE-CWM, FT: Finetune, CWM: Class Weight Masking, Tot: Total
}
\begin{scriptsize}
\label{tab:TimeCIFAR100}
\begin{tabular}
{|p{0.1\linewidth}||p{0.1\linewidth}|p{0.15\linewidth}||p{0.15\linewidth}|p{0.15\linewidth}|p{0.15\linewidth}|p{0.12\linewidth}|}
\hline
\multirow{2}{*}{\textbf{Model}}& \multirow{2}{*}{\textbf{Method}}&\multirow{2}{*}{\textbf{Component}}&\multicolumn{4}{c|}{\textbf{Forget Classes}(Time) $\downarrow$}\\
\cline{4-7} 
&&&$\mathcal{F}_1$&$\mathcal{F}_2$&$\mathcal{F}_3$&$\mathcal{F}_4$\\
\hline
\hline
\multirow{13}{*}{ResNet18}&\multirow{2}{*}{R}&Ave&37.8354&30.614&23.9884&16.8228\\
\cline{3-7}
&&\cellcolor{gray!50} Tot(5) &\cellcolor{gray!50} 189.177 & \cellcolor{gray!50} 153.07 & \cellcolor{gray!50} 119.942 & \cellcolor{gray!50}84.114\\
\hhline{|~||======}
&\multirow{3}{*}{OSR}&Pro&0.162&0.358&0.353&0.721\\
\cline{3-7}
&&Red&0.031&0.029&0.025&0.026\\
\cline{3-7}
&&\cellcolor{gray!50}Total&\cellcolor{gray!50}\textbf{0.193}&\cellcolor{gray!50}\textbf{0.387}&\cellcolor{gray!50}\textbf{0.378}&\cellcolor{gray!50}\textbf{0.747}\\
\hhline{|~||======}
&\multirow{5}{*}{\cite{TCMK23}}&NG&3.784&61.691&124.666&187.689\\
\cline{3-7}
&&RC&6.948&6.461&7.177&6.841\\
\cline{3-7}
&&Im&199.902&1156.747&257.804&315.584\\
\cline{3-7}
&&Re&12.765&107.521&17.752&3.415\\
\cline{3-7}
&&\cellcolor{gray!50}Tot&\cellcolor{gray!50}223.399&\cellcolor{gray!50}1332.42&\cellcolor{gray!50}407.399&\cellcolor{gray!50}513.529\\
\hhline{|~||======}
&\multirow{3}{*}{\cite{nature_feature}}&FT&3.577&9.966&15.513&18.982\\
\cline{3-7}
&&CWM&0.0&0.001&0.001&0.002\\
\cline{3-7}
&&\cellcolor{gray!50}Tot&\cellcolor{gray!50}3.577&\cellcolor{gray!50}9.967&\cellcolor{gray!50}15.514&\cellcolor{gray!50}18.984\\
\hhline{|=||======}
\multirow{13}{1cm}{MobileNet V2}&\multirow{2}{*}{R}&Ave&57.3078&47.0098&36.2396&25.5374\\
\cline{3-7}
&&\cellcolor{gray!50} Tot(5) &\cellcolor{gray!50} 286.539 & \cellcolor{gray!50} 235.049 & \cellcolor{gray!50} 181.198 & \cellcolor{gray!50} 127.687\\
\hhline{|~||======}
&\multirow{3}{*}{OSR}&Pro&0.164&0.429&0.421&0.402\\
\cline{3-7}
&&Red&0.03&0.041&0.056&0.048\\
\cline{3-7}
&&\cellcolor{gray!50}Total&\cellcolor{gray!50}\textbf{0.194}&\cellcolor{gray!50}\textbf{0.47}&\cellcolor{gray!50}\textbf{0.477}&\cellcolor{gray!50}\textbf{0.45}\\
\hhline{|~||======}
&\multirow{5}{*}{\cite{TCMK23}}&NG&5.539&97.707&194.632&294.216\\
\cline{3-7}
&&RC&7.031&6.921&6.561&7.338\\
\cline{3-7}
&&Im&221.252&782.43&646.702&905.26\\
\cline{3-7}
&&Re&77.738&82.331&7.078&4.546\\
\cline{3-7}
&&\cellcolor{gray!50}Tot&\cellcolor{gray!50}311.56&\cellcolor{gray!50}969.389&\cellcolor{gray!50}854.973&\cellcolor{gray!50}1211.36\\
\hhline{|~||======}
&\multirow{3}{*}{\cite{nature_feature}}&FT&3.086&8.925&15.918&19.688\\
\cline{3-7}
&&CWM&0.001&0.001&0.001&0.002\\
\cline{3-7}
&&\cellcolor{gray!50}Tot&\cellcolor{gray!50}3.087&\cellcolor{gray!50}8.926&\cellcolor{gray!50}15.919&\cellcolor{gray!50}19.688\\
\hline
\end{tabular}
\end{scriptsize}
\centering
\end{table}

\subsubsection{Coverage} 
Next we consider the coverage metric. Recall that this provides the average rank of the true label in the output of the label removal process.     In our experiment, since the number of removed labels grows, in order to allow for comparison to be done between settings with different size of labels, we consider a normalized coverage, $\overline{Cov}_\mathcal{R}$ which divides $Cov_{\mathcal{R}}$ by the number of possible labels the resulting model considers. Hence, this divides the coverage of Base Model, UNSIR, and NHLE-CWM by the original number of labels (10 for CIFAR-10, 100 for CIFAR-100 and VGGFace-100) while the coverage of retrained and OSR approach is divided by a smaller number since the output model no longer considers the removed labels. The summary of the result can be found in Table \ref{tab:CM}.

As observed, OSR outperforms even the retrained model in terms of coverage where for retained labels, the true label is constantly predicted towards the top of the list while removed labels are never considered. The performance of OSR becomes significantly better when considering larger set of labels. This suggests that OSR constantly puts the true label towards the top of the list and the larger set of possible labels causes this to have a much smaller normalized coverage.

\subsubsection{KL Divergence}
We next consider the KL divergence comparison. Recall that since our aim is to approximate the model obtained through retraining, in this section, we only consider the KL divergence between retraining and OSR as well as the two state-of-the-art solutions, UNSIR \cite{TCMK23} and NHLE-CWM \cite{nature_feature}. We consider the average KL divergence of the outputs in two groups: when the input belongs to one of the non-removed labels $KL^{R}_{\mathcal{R}}$ and when the input belongs to one of the removed labels $KL^R_{\mathcal{F}}.$ The summary of the result can be found in Table \ref{tab:KLRM}.

As can be observed, OSR only produces output with the highest similarity with the target output from the retrained model, which shows its capability to approximate 
the behavior of a freshly retrained model. It can also be observed that with the bigger number of removed labels, OSR exhibits much slower performance degradation rate in approximating the output of the retrained model, providing evidence of its robustness in handling multiple labels removal.

We further report the summary of KL divergence measurement between the solutions and the base model, which are denoted by $KL^B_{\mathcal{R}}$ and $KL^B_{\mathcal{F}}$ respectively and can be found in Table  \ref{tab:KLBM}.

As can be observed, OSR consistently produces outputs that are similar to that of the base model, which shows that the amount of knowledge being removed is due to the label removal. This makes sense considering that OSR only produces a filter to modify the output without changing the model itself. This also explains the fact that OSR output still exhibits similar behaviour to the base model when fed a data point from the removed label.

\subsubsection{Processing Time}
Lastly, we compare the time required for solutions of label removal, namely, retraining, OSR, UNSIR \cite{TCMK23}, and NHLE-CWM \cite{nature_feature}. As previously discussed, the time measurement for the different approaches are decomposed to main components while only the total time is being compared. The summary of the time measurement for datasets CIFAR10, CIFAR100, and VGGFace-100 can be found in Tables \ref{tab:TimeCIFAR10}, \ref{tab:TimeCIFAR100}, and \ref{tab:TimeVGGFace100} respectively.

It is easy to observe that in all cases, the retraining method, UNSIR, and NHLE-CWM are finetuning methods while OSR is an output processing filter, causing the processing time for OSR to be significantly lower than any of the three other solutions.

\section{Conclusion}
Label removal is a common challenge when label taxonomies evolve and categories must be updated or discarded. Most existing methods suffer from several limitations: reliance on access to the original data, high computational and storage costs, inconsistent results, limited scalability, and potential degradation of model utility. To this end, we proposed a novel paradigm that redistributes probability mass in the output space to approximate the post-removal behavior of a retrained model. Viewed as a modular output-filter, it avoids any feature-space adjustments or loss-function convergence, addressing scalability concerns. Because it operates only on the remaining labels and the model’s prior output confidences, it also mitigates privacy issues typical of data-centric solutions. Extensive experiments across multiple classification tasks demonstrate competitive performance to full retraining, with notable gains in efficiency and privacy preservation.


\subsubsection{Disclosure of Interest}
The authors declare that they have no conflict of interest.

\bibliographystyle{splncs04}
\bibliography{ref}

%




\end{document}